\documentclass[letterpaper, 10 pt, conference]{ieeeconf}  

\IEEEoverridecommandlockouts                              

\usepackage{tikz}
\usepackage{amsmath}
\usepackage[ruled, lined,commentsnumbered]{algorithm2e}
\usepackage{amssymb}
\usepackage{graphicx}
\usepackage{subfigure}
\usepackage{booktabs}
\title{\LARGE \bf
Steadily Learn to Drive with Virtual Memory
}

\author{Yuhang Zhang$^{1}$, Yao Mu$^{1}$, Yujie Yang$^{1}$, Yang Guan$^{1}$, Shengbo Eben Li$^{1 \star}$, Qi Sun$^{1}$ and Jianyu Chen$^{2}$ 
\thanks{*  This study is supported by International Science \& Technology Cooperation Program of China under 2019YFE0100200, Tsinghua University-Toyota Joint Research Center for AI Technology of Automated Vehicle and Tsinghua University-Didi Joint Research Center for Future Mobility. }

\thanks{$^{1}$Yuhang Zhang, Yao Mu, Yujie Yang, Yang Guan, Shengbo Eben Li, Qi Sun are with State Key Lab of Automotive Safety and Energy, School of Vehicle
and Mobility, Tsinghua University, Beijing, 100084, China. All correspondence should be sent to S. Eben Li with email 
        {\tt\small lisb04@gmail.com}}

\thanks{$^{2}$Jianyu Chen is with   Institute of Interdiscriplinary Information Sciences, Tsinghua University, Beijing, China and Shanghai Qi Zhi Institute {\tt\small jianyuchen@tsinghua.edu.cn}}%
}

\begin{document}

\maketitle
\thispagestyle{empty}
\pagestyle{empty}

\begin{abstract}
    Reinforcement learning has shown great potential in developing high-level autonomous driving. However, for high-dimensional tasks, current RL methods suffer from low data efficiency and oscillation in the training process. This paper proposes an algorithm called Learn to drive with Virtual Memory (LVM) to overcome these problems. LVM compresses the high-dimensional information into compact latent states and learns a latent dynamic model to summarize the agent's experience. Various imagined latent trajectories are generated as virtual memory by the latent dynamic model.  The policy is learned by propagating gradient through the learned latent model with the imagined latent trajectories and thus leads to high data efficiency. Furthermore, a double critic structure is designed to reduce the oscillation during the training process. The effectiveness of LVM is demonstrated by an image-input autonomous driving task, in which LVM outperforms the existing method in terms of data efficiency, learning stability, and control performance. 

\end{abstract}

\section{Introduction}
    \par With the development of artificial intelligence technologies, autonomous driving has become an important tendency in the automotive industry for its potential to improve road safety, reduce fuel consumption, and improve traffic efficiency \cite{9046805, bimbraw2015autonomous}. There are two kinds of schemes widely employed by the autonomous driving system: hierarchical scheme \cite{Urmson2008Autonomous, montemerlo_junior_2008, duanHierarchicalReinforcementLearning2020, muMixedActorCriticEfficient} and end-to-end scheme \cite{jaritz_end--end_2018, pomerleau_alvinn_1988,bojarski_end_2016}. The hierarchical scheme divides the entire autonomous driving system into several modules, including environment perception, decision-making, and motion control \cite{Urmson2008Autonomous, montemerlo_junior_2008, 6875912}. Each module needs to be manually designed and tuned to achieve satisfying driving performance. Moreover, these modules are hierarchically structured, making errors easily propagated to downstream modules. End-to-end methods learn a single driving module which directly generates control commands from raw sensor inputs. It has gained more and more attention recent years due to its straightforward system design and better overall performance compared with its hierarchical counterparts theoretically \cite{bojarski_end_2016}.
    
    In the end-to-end scheme, Reinforcement learning (RL) has great advantages over supervised learning because of the lack of natural driving data with such high-dimensional inputs \cite{jaritz2018end,chen2020interpretable, chen2020stabilization}. It learns from interactions with an environment or an analytical model, achieving incredible success in several fields vary from games to robotics  \cite{mnih_human-level_2015,vinyals_grandmaster_2019}. 
     Yu et al. (2016) trained an agent to learn the ability of turning operation and navigation in the game JavaScript Racer by Deep Q-Network (DQN) \cite{yu_deep_nodate}.
     Jaritz (2018) utilized Asynchronous Advantage Actor-Critic (A3C) to learn the vehicle control in a physically and graphically realistic rally game in an end-to-end manner \cite{jaritz_end--end_2018}.
     Kendall et al. (2018) trained a Deep Deterministic Policy Gradient (DDPG) controller in the lane-keeping task, in which a single front-view image was used as input  \cite{kendall_learning_2018}. 
     However, when applied in real autonomous driving scenarios, these methods still suffer from low data efficiency and final performance. 
     
     \par Several studies in RL have been proposed to deal with high-dimensional sensory inputs. Ha et al. (2018) proposed World Model, which learned a variational auto-encoder (VAE) to extract low-dimensional latent features of the image input \cite{ha_world_2018}. 
     The VAE \cite{kingma_auto-encoding_2014,higgins_vae_2017} learns the compact latent representation by variational inference. 
     Then an RNN-based prediction model is learned to predict the state transition in the latent space independently from the representation model and thus limits the prediction accuracy and the final policy performance.
     In order to learn a better representation model and prediction model, Hafner et al. (2018) proposed to use Recurrent Stochastic State Model (RSSM) to represent the environment dynamics  \cite{hafner_learning_nodate}.
     RSSM model trained the representation and prediction model together by variational inference, achieving better performance than World Model. RSSM employs both stochastic variables and deterministic variables, which has the ability of uncertainty representation and achieves higher predicting accuracy. Lee et al. (2020) presented SLAC, which used a latent dynamic model with only stochastic representation and achieved comparative performance of algorithms with RSSM  \cite{lee_stochastic_2020}. After mapping the high-dimensional images into the low-dimensional latent space, the performance of RL algorithms in image-input task are improved effectively compared with those standard RL methods.
     
     However, when applied in autonomous driving scenarios environment, these methods still suffer from low data efficiency and oscillation in the training process. The interaction data collected by the exploration in the real physical world is expensive. Moreover, the unstable policy during the training process may lead to safety accidents and make the training process dangerous.
     In 2020, Hafner et al. proposed a novel model-based algorithm with latent representation, Dreamer, to improve data efficiency  \cite{hafner_dream_2020}. 
     Nevertheless, generating a single latent imagined trajectory from each starting point limits the full use of latent dynamic model, which leaves considerable room for improvement in data efficiency.    Meanwhile, the overestimation of the state value caused by the model error leads to unstable policy during the training process.
     Hasselt et al. (2015) argued that any estimation errors caused by system noise, function approximation, etc., will induce an upward bias of the values \cite{van_hasselt_deep_2015}.  In Dreamer, the error of the learned latent dynamic model results in overestimation of the state value and unstable policy. 
     
	To overcome the aforementioned challenges, we propose a novel actor-critic algorithm called \textbf{L}earn to drive with \textbf{V}irtual \textbf{M}emory (LVM). The LVM learns a virtual latent environment model from real interaction data of the agent's past experience to predict the environment transition dynamics.
	The agent explores the virtual environment and records the latent imagined trajectories as the virtual memory. The policy is then optimized by the virtual memory and does not need real interaction data, significantly improving the data efficiency. Inspired by Clipped Double-Q algorithm  in model-free methods \cite{fujimoto_addressing_2018, haarnoja_soft_2018,duan_distributional_2020}, LVM learns two critics independently to reduce the overestimation of the state value and stabilize the training process of the model-based RL.
	
	The contributions of this paper are as follows,
	\begin{itemize}
	\item An model-based RL algorithm called \textbf{L}earn to drive with \textbf{V}irtual \textbf{Me}mory (\textbf{LVM}) is proposed to improve the data efficiency in the driving policy training process;
	\item A double critic structure of state value estimation is designed for model-based RL to make the training process more stable than its counterparts by reducing the value overestimation caused by the model errors and noise.
	\end{itemize}
	\par This paper is organized as follows. Section II describes the autonomous driving problem. In section III, the proposed algorithm, LVM, is presented. Section IV discusses the experiment and training results. Section V concludes this paper.

\section{Preliminaries}
    \par Nowadays, self-driving vehicles employ a significant number of expensive sensors and high energy-consuming chips, such as LiDAR and GPUs, to achieve high-level autonomous abilities. 
    Here, we try to build an autonomous system with a single cheap camera.
    Historically, researchers investigate the end-to-end autonomous driving systems, which use camera image as input \cite{yu_deep_nodate,jaritz_end--end_2018,kendall_learning_2018}.
    However, these methods suffer from the low training efficiency due to high input dimensions.
    Different from aforementioned methods, the autonomous driving task is described as a Partial Observable Markov decision Process (POMDP)  \cite{Shengbo2019} with high dimensional observation $o\in \Omega$, hidden state $s\in \mathcal{S}$, action $a\in \mathcal{A}$, transition model $P(s^\prime |s, a): \mathcal{S} \times \mathcal{A} \rightarrow \mathcal{P}(s')$, reward function $r(s, a):\mathcal{S} \times \mathcal{A}\rightarrow \mathbb{R}$ and observation function $O(o|s^\prime, a):\mathcal{S} \times \mathcal{A}\rightarrow \Omega$. 
	
	\par In this problem, the observation $o$ is a three-channel RGB image generated by the camera attached to the front part of the vehicle. 
	The state $s$ represents the dynamic state of the vehicle and surrounding environment, such as lane position and obstacle position. 
	The state is learned by a representation learning algorithm, in which its ground-truth is not accessible. 
	The transition model $P$ determines the dynamics of the state given the current state and action. 
	The action $a$ is given by the policy $\pi (a|o_{\le t}, a_{<t})$ which is a mapping from the direct product space of $\mathcal{S}$ and $ \mathcal{A}$ to action space $ \mathcal{A}$. 
    The subscript ${\le t}$ stands for all the information before or at time step $t$,  while ${< t}$ does not include time step $t$. 
	The definition of policy function is different from that of MDP because of the lack of Markov property in the observation space. 
    Reward function $r$ is given by the environment, which measures how good an action is. 
    Good actions, such as getting closer to the centerline of the road, lead to higher rewards, while bad actions, such as crashing out of the road, lead to lower reward.
    At each step, the agent takes an action based on historical observations and actions, receives a reward $r$, and arrives at a new state $s^\prime$. 
    However, the agent cannot get access to the new state and only receives a new observation $o^\prime$ generated by the observation function.
    
    The goal of the agent is to find the optimal policy parameter $\theta$ that maximizes the long-horizon accumulated rewards represented by a value function,
    \begin{align}
        V_\pi(s)=\mathbb{E}_\pi \left\{ \sum_{t=0}^\infty \gamma^tr_t \left| \right. s_0 = s\right\}.
    \end{align}

    However, such RL algorithms with high-dimensional observation space will result in quite inefficient training. 
    Recently studies have shown the potential of learning a low-dimensional latent dynamic model in such problems.
    The latent dynamic model aims to extract useful information from raw data and represents it in a compact latent state.
    
\section{Learn to Driver with virtual memory}
\begin{figure}
	\centering
	\includegraphics[width=0.425\textwidth]{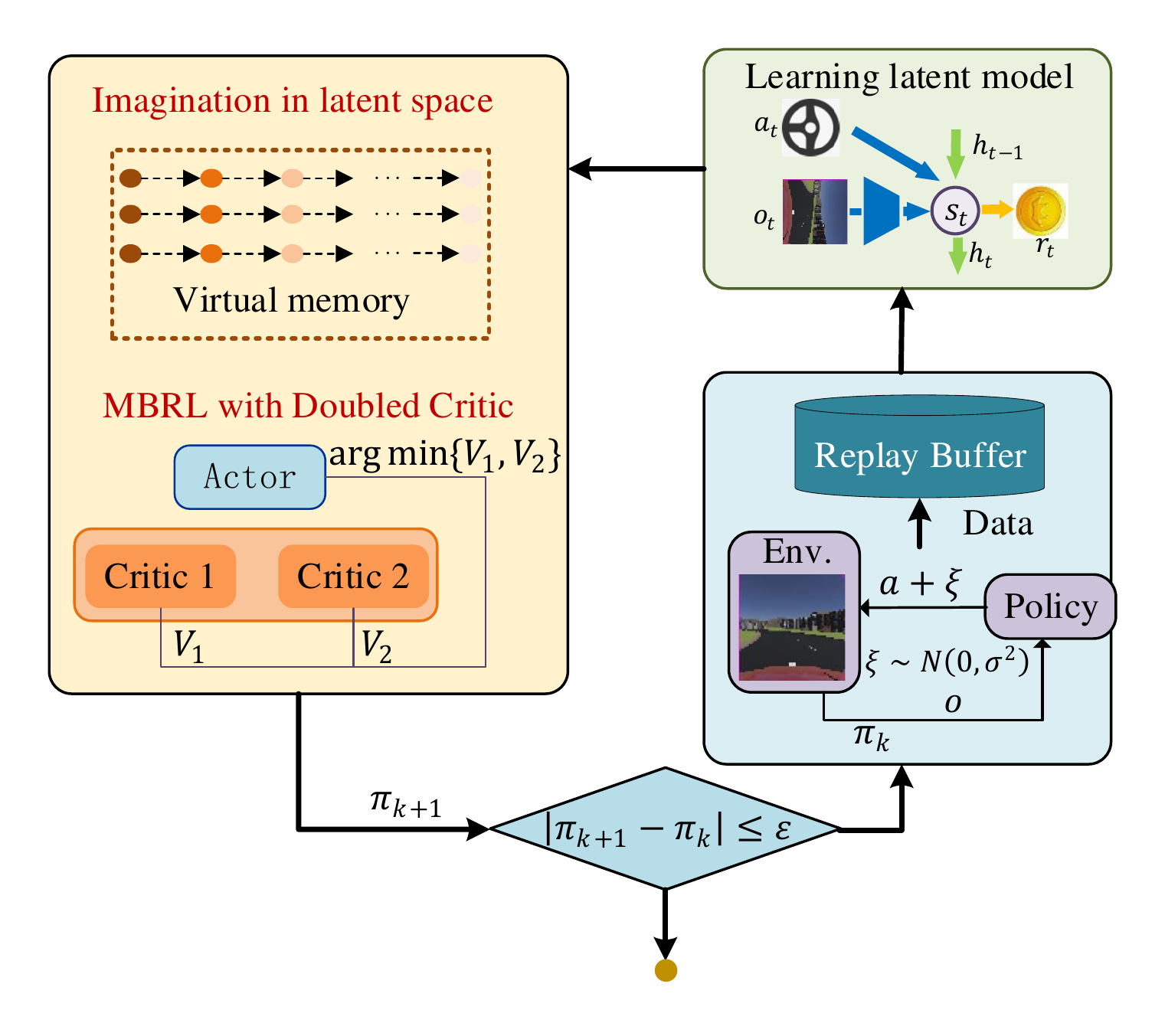}
	\caption{Framework of LVM}
	\label{fig:framework}
\end{figure}

As shown in Fig. \ref{fig:framework}, the training iteration of LVM has three alternate steps, the latent dynamic model learning, model-based RL training in latent space, and data collecting. In the first step, the latent dynamic model compresses the high-dimensional inputs into the low-dimensional latent states and approximates the environment dynamic in the learned latent space. In the second step, the latent dynamic model is utilized to generate virtual trajectories by imagination. The policy and value function is then updated by these virtual trajectories. In the third step, new interaction data is sampled from the real environment by the updated policy to enrich the replay buffer.

\subsection{Latent dynamic model for autonomous driving}
\begin{figure}
		\centering
		\includegraphics[width=0.45\textwidth]{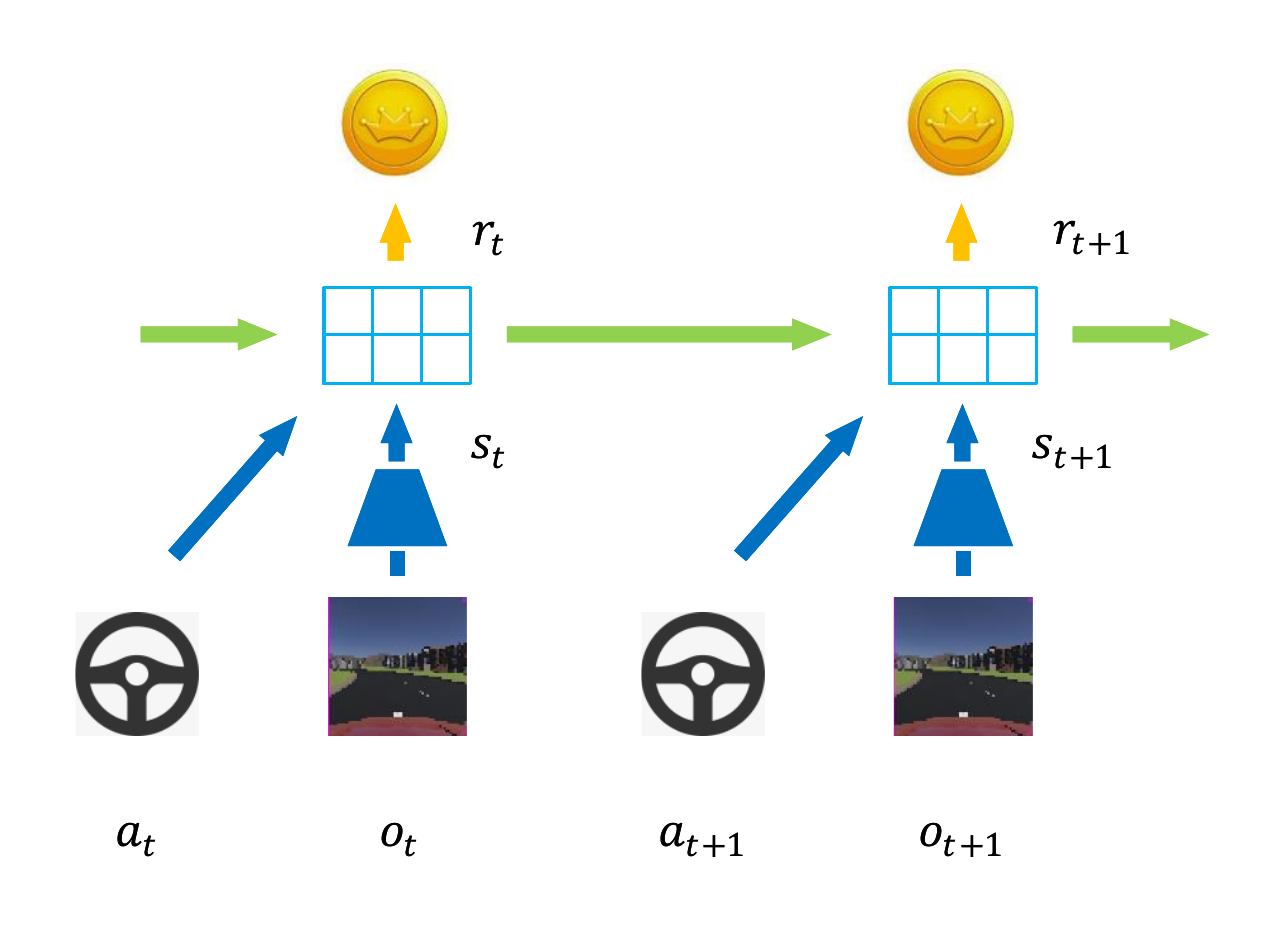}
		\caption{Latent dynamic model}
		\label{fig:latent_dynamic_model}
	\end{figure}

	As show in Fig. \ref{fig:latent_dynamic_model}, latent dynamic model consists of two basic elements, the encoder and the latent transition model. The encoder maps the observation $o$ into a latent state $s$. Then, the latent transition model define the dynamic of the latent state, i.e. $s^\prime = f(s, a)$.
	
	In this paper, recurrent state space model (RSSM) learned from the interaction data is used to approximate the environment dynamic effectively. 
	RSSM model encodes the world into two kinds of states, the stochastic state $s$ and the deterministic state $h$. 
	The usage of the stochastic state makes the model able to represent the uncertainty of the environment. 
	The deterministic state effectively improves the predicting accuracy of the model. 
	Hafner et al. (2020) argued that both stochastic part and deterministic part contribute to the performance of latent dynamic model \cite{hafner_learning_nodate}.
	
	The RSSM has four main components: 
	
	\begin{minipage}{\linewidth}
	    \centering
	    \begin{minipage}[t]{0.48\linewidth}
	   \begin{equation*}
	        \begin{split}
        		&\text{Deterministic state model:}\\
        		&\text{Stochastic state model:}\\
        		&\text{Reward model:}\\
        		&\text{Observation model:}
		    \end{split}
		\end{equation*}
	    \end{minipage}
	    \hfill
	    \begin{minipage}[t]{0.48\linewidth}
	    \begin{equation}
	        \begin{split}
		h_t &= f_\psi(s_{t-1}, a_{t-1}, h_{t-1})\\
		s_t &\sim p_\psi(s_t |h_{t})\\
		r_t &\sim p_\psi(r_t|h_t, s_t)\\
		o_t &\sim p_\psi(o_t|h_t, s_t),
		\end{split}
		\end{equation}
	    \end{minipage}
	
	\end{minipage}
	where $\psi$ is the parameter of the RSSM model. The deterministic state model and stochastic model together form the representation model, which maps the high dimensional observation space into a low dimensional feature space, which is called the latent space. In the latent space, it is easier to extract useful information for the subsequent policy training. The deterministic state model is usually approximated by recurrent neural networks such as Long Short-Term Memory (LSTM) \cite{1997Long} and  Gated Recurrent Unit (GRU)  \cite{chung2014empirical}, and thus $h_t$ is the hidden variable which transmits historical information between different time steps. The stochastic state model predicts the distribution of the latent state according to the hidden state. Meanwhile, the two state models also compose the transition model of the latent dynamics, which can be used to rollout virtual trajectories and improve the policy. The policy training process does not interact with the real environment, thus a reward model is essential for the policy improvement step and policy evaluation step of the RL algorithm. The observation is an essential part when training the representation model in variational inference method.
	
	The RSSM model is optimized by variational inference method, in which the evidence lower bound (ELBO) is maximized. The loss function is defined as 
	\begin{equation}
		\begin{split}
		J_{\text{RSSM}} &= \mathbb{E}_p\left(\sum_t J_o + J_r + J_D \right),\\
		\text{with}\quad J_o &= \ln p_\psi(o_t|s_t)\\
		J_r &= \ln p_\psi(r_t|s_t)\\
		J_D &= - \operatorname{KL}\left(p_\psi(s_t | s_{t-1}, a_{t-1}, o_t)||q_\psi(s_t|s_{t-1}, a_{t-1})\right).
		\end{split}
	\end{equation}
	
	The RSSM loss consists of three parts. $J_o$, $J_r$ is the log-likelihood of the observation and reward, which can also be regarded as the reconstruction loss. In the implementation, the distributions of observation and reward are considered as Gaussian variables of which the mean is the output of the observation model and reward model, and the variance is a predefined constant. Meanwhile, $J_D$ is the state prediction loss that minimizes the KL divergence between the prior distribution and posterior distribution of state transition, which is a metric for the prediction accuracy of RSSM. Because the agent has no access to the real environment, accurate prediction is necessary for the policy training.
	
\subsection{Policy optimization with virtual memory} 
\subsubsection{Virtual memory imagination}
\begin{figure}
		\centering
		\includegraphics[width=0.5\textwidth]{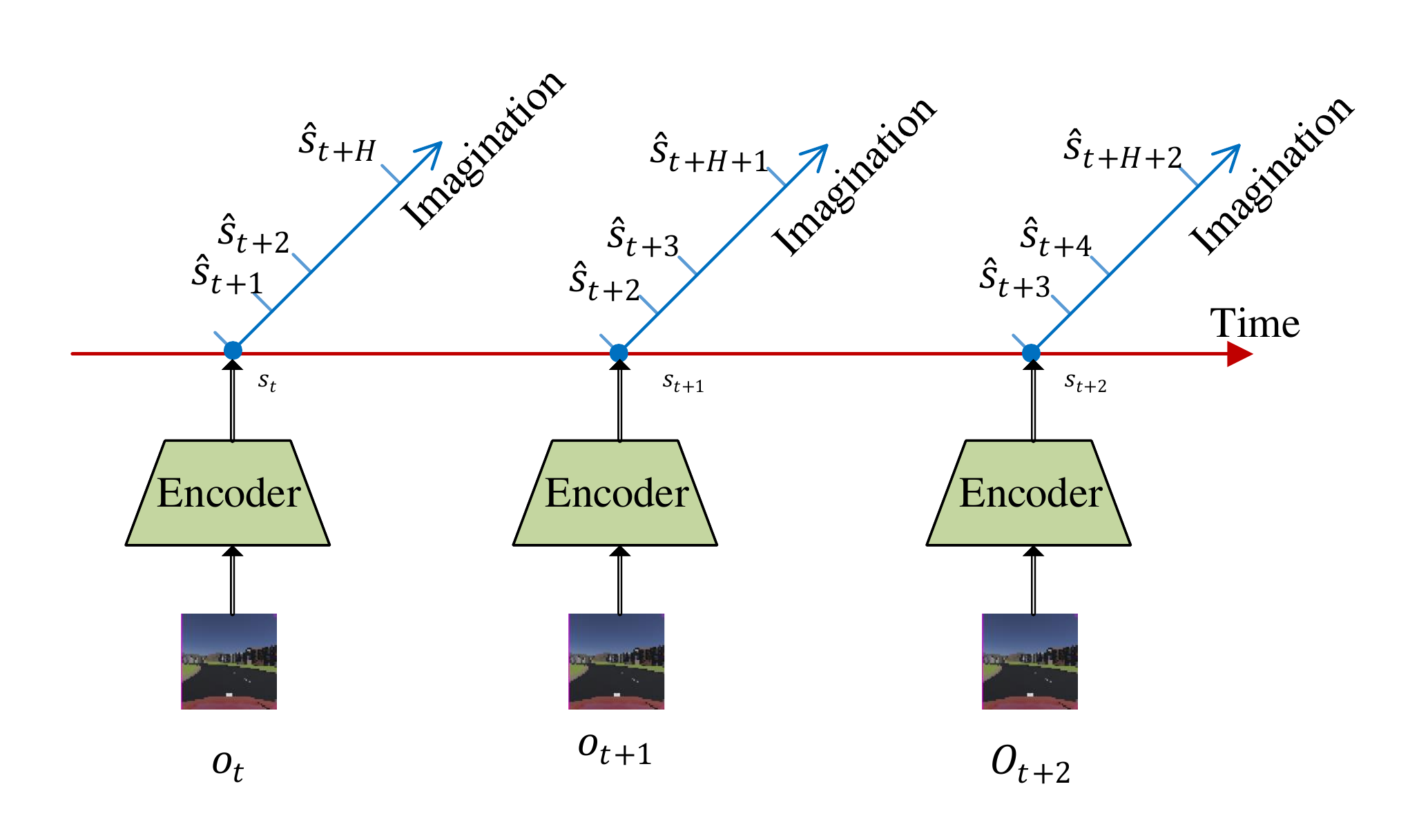}
		\caption{Virtual trajectory imagination}
		\label{fig:imagination}
	\end{figure}

	After obtaining the RSSM model, it can be used to generate trajectories $\hat{\tau}_i$ in virtual memory. At first, several real trajectories are sampled from the database and serve as the virtual trajectory's starting points. Then, as shown in Fig.  \ref{fig:imagination}, the RSSM model is used to generate trajectories of a certain length from these states. Due to the stochastic property of the latent dynamic model, various trajectories are generated from each starting point. These trajectories are used to update the value function and policy function.
	Interacting with the learned RSSM model is more efficient than that of real environment. Therefore, the more stable on-policy method is naturally used instead of its off-policy counterparts, which is commonly used in model-free RL to increase data efficiency.
	
\subsubsection{Policy optimization with virtual memory}
	In policy optimization, the latent state $s_t$ and hidden state $h_t$ are concatenated to form the policy's real input. 
	The latent state $s_t$ stands for the extracted information of current observation. 
	Meanwhile, hidden state $h_t$ represents the historical information, which is necessary because of the POMDP property. The policy is trained in an actor-critic architecture. 
	The actor outputs action selected by the agent according to current state. And the critic approximates the value function of the states.
	\begin{align}
	a_t &= \pi_\theta (s_t, h_t),\\
	V_\phi(s_t, h_t) &= \mathbb{E}_{\pi_\theta}(\sum_{\tau=t}^H\gamma^{\tau - t}r_\tau),
	\end{align}
	where $\theta, \phi$ are the parameters of the approximated policy and value.
	
	The training process can be divided into two alternate processes. First, the critic is optimized by TD updating. Considering that the training data is generated by the learned RSSM model, which may lead to high variance, $\operatorname{TD}(\lambda)$ trick is used to reduce the variance of policy evaluation. $\operatorname{TD}(\lambda)$ return, $V_\lambda(s_t, h_t)$, is defined as
	\begin{equation}
	V_\lambda(s_t, h_t) = (1-\lambda)\sum_{n=1}^{H-1} \lambda^{n-1} V_N^n(s_t, h_t) + \lambda^{H-1}  V_N^H(s_t, h_t)
	\end{equation}
$$ \text{where } \small{V_N^k(s_t, h_t) = \mathbb{E}_{\pi_\theta}\left\{\sum_{\tau=t}^{t+k -1}\gamma^{\tau - t}r_\tau + \gamma^{k} V_\phi(s_{t+k}, h_{t+k})\right\}.}$$
	
	Inspired by Clipped Double Q-learning trick, two value function, $V^{(1)}(s, h;\phi _1)$ and  $V^{(2)}(s, h;\phi _2)$ is learned to reduce overestimation and stabilized the training process. When estimating the target, a minimum operation is utilized, i.e.
	\begin{equation}
	    V_{\text{target}}(s_t, h_t) = \min \left\{V_\lambda^{(1)}(s_t, h_t;\phi _1), V_\lambda^{(2)}(s_t, h_t;\phi _2)\right \}.
	\end{equation}
	The two critic is updated independently as 
	\begin{equation}
	\begin{split}
	\min_{\phi_1} J_{V_1}= \mathbb{E}_{\pi_\theta}\left( \sum_{\tau=t}^{t+H} \frac{1}{2} \lVert V_{\phi_1}(s_\tau, h_\tau) - V_{\text{target}}(s_\tau, h_\tau) \rVert^2 \right), \\
	\min_{\phi_2} J_{V_2}=\mathbb{E}_{\pi_\theta}\left( \sum_{\tau=t}^{t+H} \frac{1}{2} \Vert V_{\phi_2}(s_\tau, h_\tau) - V_{\text{target}}(s_\tau, h_\tau) \Vert^2 \right).
	\end{split}
	\end{equation}
	The actor aims to maximize the expected return of the policy which is estimated by the critic. Thus, its objective is
	\begin{equation}
		\max_\theta J_\pi =  \mathbb{E}_{\pi_\theta}\left( \sum_{\tau=t}^{t+H} V_{\text{target}}(s_\tau, h_\tau) \right) 
	\end{equation}
	
	\subsection{Data collection}
	In each iteration, new trajectories sampled by current policy is added to the replay buffer. In the sampling process, in order to fully explore the whole world, a fixed noise is attached to the policy outputs, i.e.
	\begin{align}
	    a = \pi_\theta (s_t, h_t) + \xi,\\
	   \text{with } \xi \sim N(0, \sigma^2), 
	\end{align}
	where $\sigma$ is a predefined constant. The pseudo code of LVM in shown in Algorithm \ref{alg}
	
	\begin{algorithm}[ht]
		\SetAlgoLined
		
		Initialize replay buffer $\mathcal{D}$ with $S$ episodes with random policy\\
		Initialize neural network parameters $\theta$, $\phi_1$, $\phi_2$, $\psi$ \\
		Initialize RSSM learning rate $\alpha_m$, value learning rate $\alpha_v$ and policy learning rate $\alpha_\pi$
		
		\For{e = 1\dots MaxEpoch}{
			\For{i = 1\dots TrainFreq}{
				Sample sequence batch $\{(a_t, o_t, r_t )\}_{t=t:t+L}$ from the replay buffer\\
				{\texttt{//Dynamics learning}}\\
				$\psi \leftarrow \psi + \alpha_m\nabla_\psi J_{\text{RSSM}}$\\
				{\texttt{//Policy learning}}\\
				\For{k=1\dots TrajNum}{
				Rollout trajectories $\{(s_\tau, a_\tau, r_\tau)\}_{\tau=t: t+H}$ from each $s_t$ in the sequence batch\\}
				
				$V_{\text{target}} \leftarrow \arg\min\{ V_\lambda^{(1)}, V_\lambda^{(2)}\}$\\
				$\phi_1 \leftarrow \phi_1 - \alpha_V \nabla_{\phi_1} J_{V_1}$\\
				$\phi_2 \leftarrow \phi_2 - \alpha_V \nabla_{\phi_2} J_{V_2}$\\
				$\theta \leftarrow \theta + \alpha_\pi \nabla_\theta J_\pi$\\
		
		}
		{\texttt{//Data collection}}\\
		\For{j = 1\dots DataCollectFreq}{
			$s\leftarrow $env.reset()\\
			\While{not done}{
			$s\leftarrow \pi_\theta(s)$\\
			$s, r, done$ = env.step($a$)\\
			$\mathcal{D}$.append($s, a, r, done$)	
		     }
		}
		}
		\caption{Learn to driving in virtual memory (LVM)}
		\label{alg}
	\end{algorithm}

\section{Experiments}
In this section, LVM is implemented in a lane-keeping task, where the vehicle is expected to drive along the center of the lane. First, the environment setup is introduced. Second, we describe the essential details of implementation. Finally, the evaluation results are presented  and discussed.
\subsection{Environment setup}
\begin{figure}
	\centering
	\subfigure{
		\begin{minipage}[t]{0.4\linewidth}
			\centering
			\includegraphics[width=\textwidth]{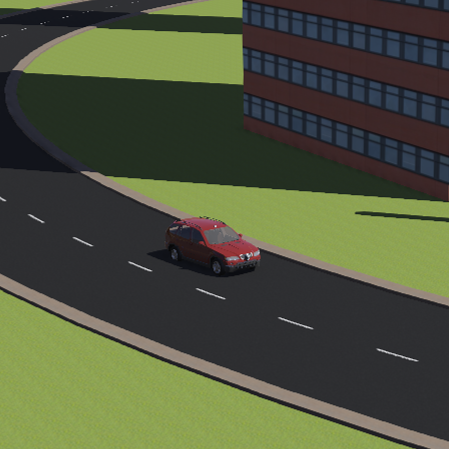}
		\end{minipage}%
	}%
	\hspace{0.1\linewidth }
	\subfigure{
		\begin{minipage}[t]{0.4\linewidth}
			\centering
			\includegraphics[width=\textwidth]{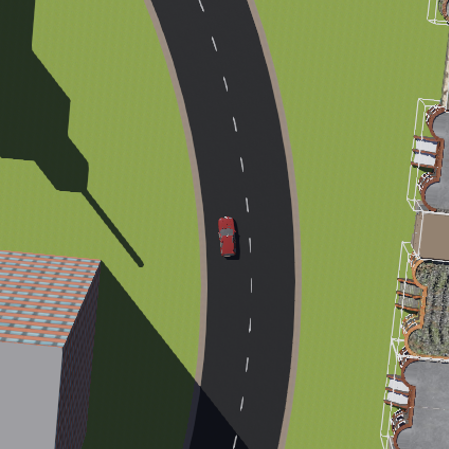}
		\end{minipage}%
	}
		
	\caption{The environment build in Webots}
	\label{fig:webots}
\end{figure}

 The high fidelity simulate environment is built in Webots, which is a professional mobile robot simulation software package \cite{Webots}. As shown in Fig. \ref{fig:webots}, the environment is a two-lane ring road with various buildings on both sides. Moreover, to present images with high fidelity, we consider light changing in the environment.

The environment takes the acceleration $a$ and steering angle $\delta$ as input. 
And the environment feedback has three elements: an image from the camera, a scalar reward and a bool variable which indicates whether the episode ends. 
The image has 3 channels with size $64\times 64$.
The underlying vehicle model used in the simulator is the bicycle model in polar coordinates \cite{rajamani2011vehicle}.

\begin{figure}
	\centering
	\includegraphics[width=0.4\textwidth]{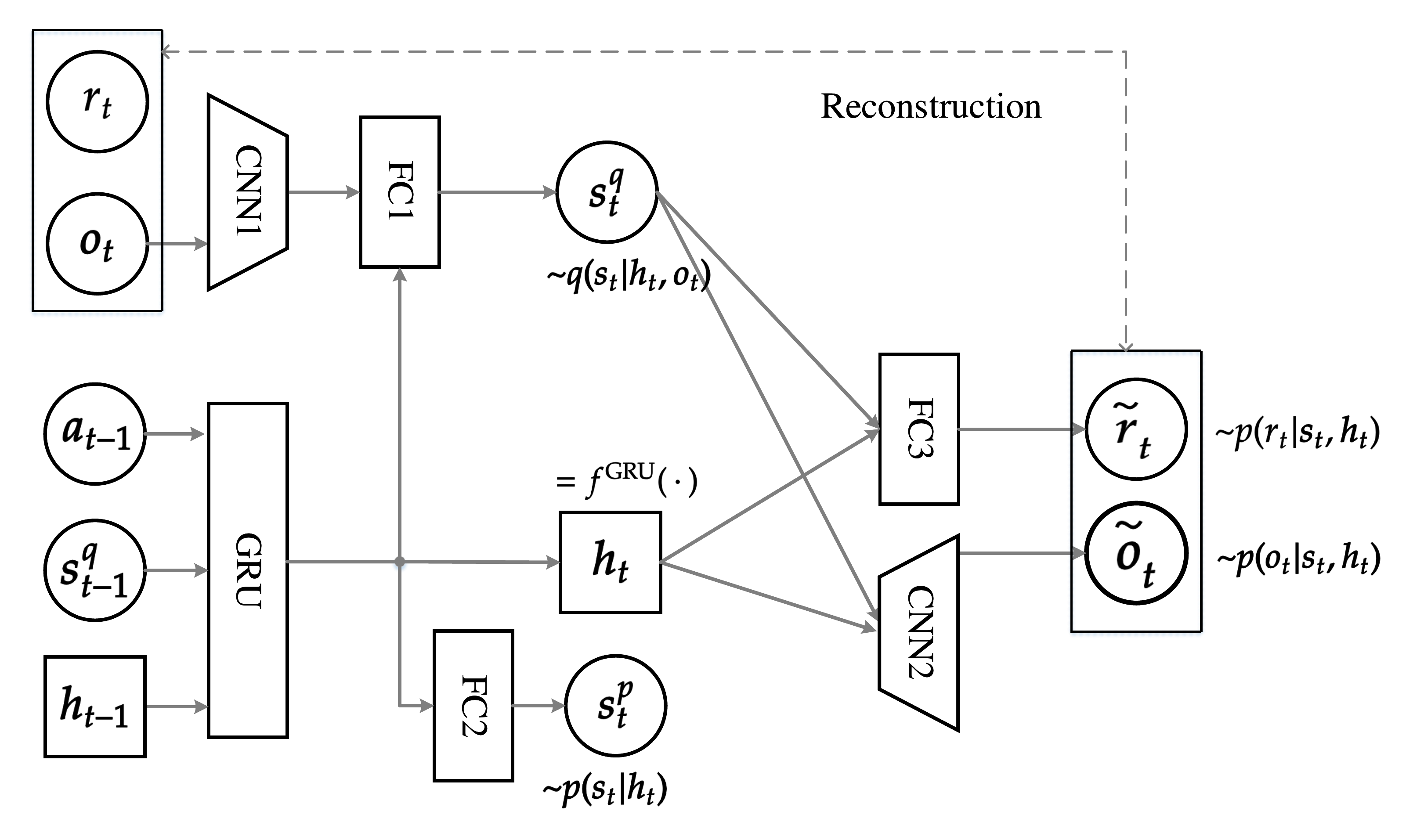}
	\caption{Recurrent state space model}
	\label{fig:rssm}
\end{figure}
In this task, the vehicle is supposed to drive in the center of the road. Meanwhile, the control variables, the acceleration and steering angle, shall be as small as possible. Therefore, the reward function is designed to penalize the deviation of the vehicle from the center line and large action outputs, 
\begin{equation}
	r = c_1 y^2 + c_2 \varphi^2 + c_3 \omega^2 + c_4\beta ^2 + c_5 (v-v_0)^2 + c_6\delta^2 + c_7 a^2
	\label{eq:reward}
\end{equation}
where $c_i$, $i=1\dots 7$ are predefined coefficients, $y$ is the distance between vehicle centroid and the road center line, $\varphi$ is the error between vehicle heading angle and the road tangent, $\beta$ is side-slip angle of the vehicle body, $\omega$ is the yaw rate of the vehicle body, $v$ stands for the longitudinal speed of the vehicle, $v_0$ is the expected longitudinal speed, $\delta$ is the steering angle, and $a$ is the longitudinal acceleration.

The design of the reward function is critical to the algorithm training. In model-free algorithms, a discrete reward is commonly used as a penalty for the agent's death, i.e., crashing out of the road. However, in model-based training, where a differential reward function is needed for the gradient computation, such a discrete reward would not work. Therefore, in this work, the reward is designed as a smooth and differential function (\ref{eq:reward}). 

\subsection{Implementation details}
\par Before LVM training starts, we pretrain RSSM to build a basic representation of the environment dynamics.
Several trajectories are sampled by a random policy, such as Gaussian policy, to build a training dataset.
Then the latent dynamic model is pretrained in the fixed dataset. After pretraining, the main part of LVM, including RSSM training, model-based RL training in latent space, and data collection, is repeated until the algorithm converges. 


The detailed structure of the RSSM model is shown as Fig. \ref{fig:rssm}, which consists of two CNNs, three fully connected networks, and a recurrent neural network. 
The policy and the value are both fully connected networks, each of which has two hidden layers with 256 units.
Key hyper-parameters of the proposed algorithm are shown in Table \ref{tab:my-table}. You can see more details in LVM's homepage \footnote{https://sites.google.com/view/reinforcement-learning-lvm}.
\begin{center}
\begin{table}[h]
\centering
\caption{Hyper-parameters}
\begin{tabular}{@{}cc@{}}
\toprule
Hyper-parameters & Value \\ \midrule
 Optimizer          &   ADMM    \\
  Replay Buffer Size         &  1e6     \\
  Batch Size  &     50  \\
  Sequence Length  &   50    \\
  Stochastic latent size  &  60     \\
  Deterministic latent size  &  256     \\
  Time step  &  0.05 s     \\
  Discounted Factor  &  0.99     \\
  Actor Learning Rate &   1e-4    \\
  Critic Learning Rate &  1e-4     \\
  RSSM learning Rate &    1e-3   \\ \bottomrule
\end{tabular}

\label{tab:my-table}
\end{table}
\end{center}

\subsection{Evaluation results}
\begin{figure*}
	\centering
    \subfigure[Real images]{
        \includegraphics[width=0.4\textwidth]{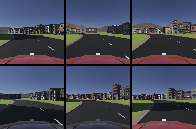}
        
    }
    \subfigure[Reconstruction images]{
	\includegraphics[width=0.4\textwidth]{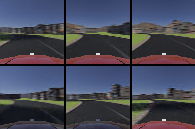}
        
    }
    \caption{The reconstruction performance of RSSM}
    \label{fig:reco}
\end{figure*}

The performance of image reconstruction is demonstrated in Fig. \ref{fig:reco}. 
The result shows that the reconstructed image contains all the information used in downstream tasks, such as the lateral position, the curvature of the road, and the lane mark, etc. Therefore, the model-based RL trained with latent dynamic model is effective in the real environment.

We compared our algorithm with Dreamer  \cite{hafner_dream_2020} and SLAC  \cite{lee_stochastic_2020}. The model-based RL algorithm, Dreamer, solves the pure image-input control task of Mujoco and achieves high final performance and data efficiency. Meanwhile, SLAC combines latent dynamic model with SAC, the state-of-the-art model-free RL algorithm, and achieves comparative performance compared with Dreamer in Mujoco.

\begin{figure}
	\centering
	\includegraphics[width=0.85\linewidth]{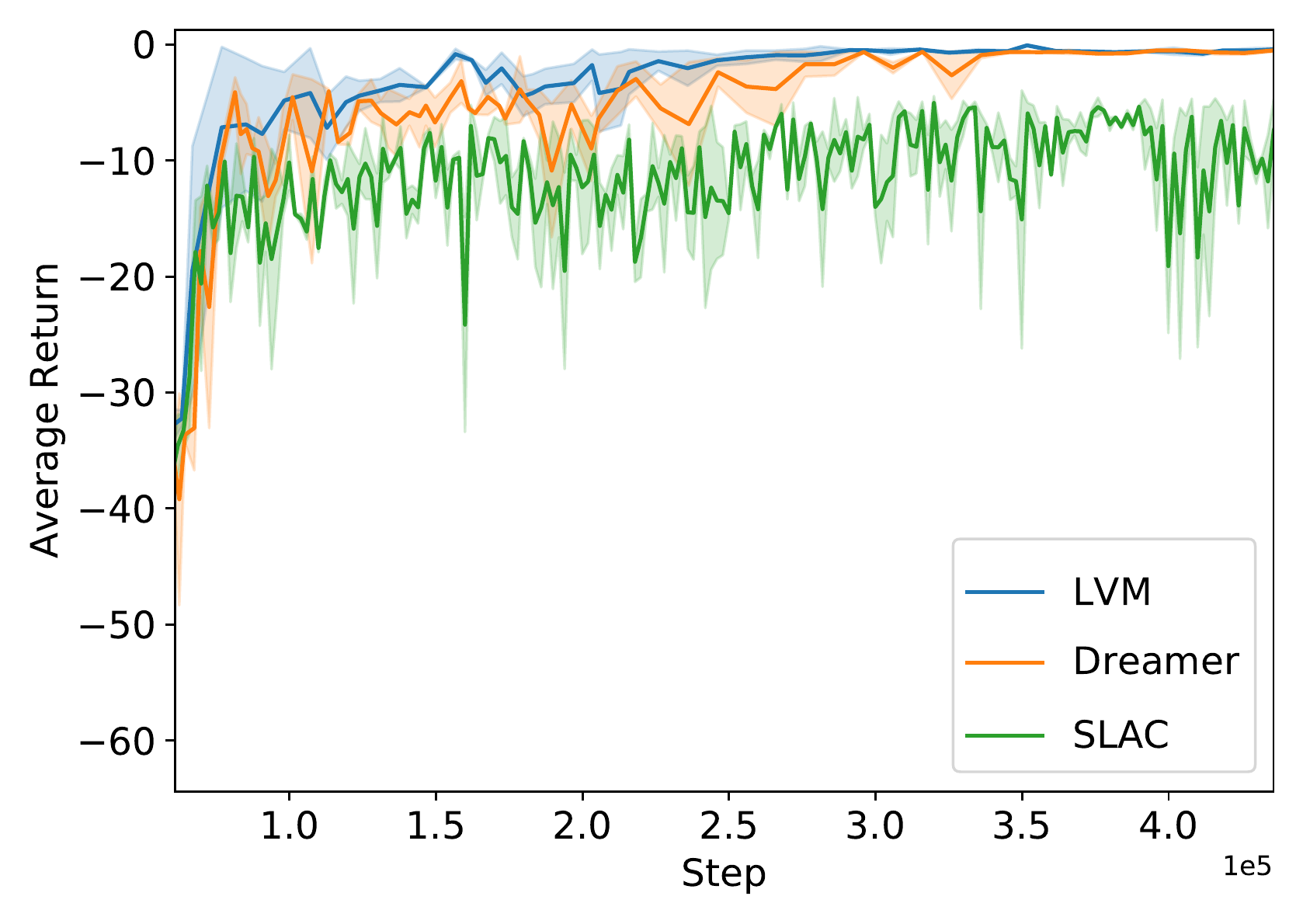}
	
	\caption{Comparison of average return}
	\label{fig:curve}
\end{figure}

In this work, the average return is utilized as the performance measure. Each algorithm is trained in five different runs with different random seeds. 
The comparison of the average return is shown in Fig. \ref{fig:curve}. 
LVM and Dreamer have better data efficiency and stability than SLAC. And their final performance apparently outperforms SLAC. Meanwhile, LVM has a more stable training process compared with Dreamer. As shown in Fig. \ref{fig:curve}, after 250k steps of training, the average return of LVM has almost converged. In contrast, the curve of Dreamer still oscillates occasionally.

\begin{figure}
	\centering
	\includegraphics[width=0.8\linewidth]{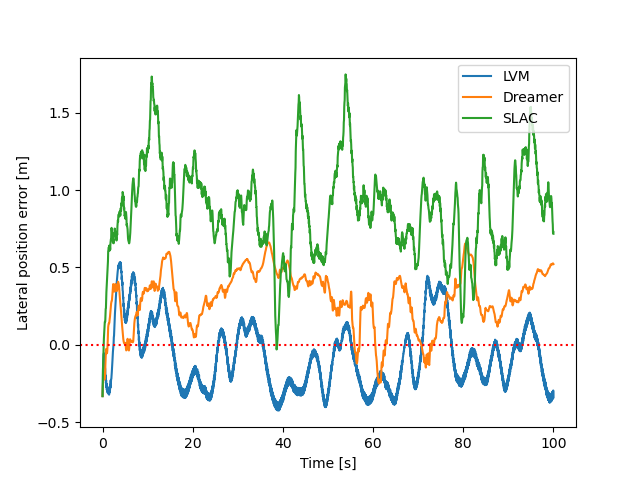}
	\caption{Comparison of control performance}
	\label{fig:lateral}
\end{figure}

The control performance of LVM and its baselines are demonstrated in Fig. \ref{fig:lateral}. 
The mean lateral position error is shown in Table \ref{tab:mean_error}. LVM has the lowest lateral position errors compared with its counterparts according to 20 random tests. The average lateral position error of LVM is $38.0 \% $ less than that of Dreamer,  $89.0 \% $ less than SLAC.

In summary, LVM exhibits the best data efficiency and convergence stability during the training process and superior control performance in the lane-keeping task.

\begin{table}[h]
\centering
\caption{Average lateral position error}
\begin{tabular}{@{}cccc@{}}
\toprule
& LVM & Dreamer & SLAC \\ \midrule
 Error [m] &0.196 & 0.316 & 0.872  \\\bottomrule
\end{tabular}
\label{tab:mean_error}
\end{table}


\section{Conclusion}
    This paper proposes a novel model-based RL algorithm called Learn to drive with Virtual Memory (LVM) to build an autonomous driving system with a front camera image as the only sensor. LVM learns a latent dynamic model, which compresses high-dimensional raw data into a low-dimensional latent space and predicts the environment transition in the latent space. Instead of interacting with the real environment, LVM learns the optimal policy by virtual memory generated by the learned latent dynamic model, which improves the data efficiency significantly.
    In the process of policy learning, the double critic structure is designed to stabilize the algorithm training and reduce the oscillation occurrence. The performance of the algorithms is evaluated in the lane-keeping task in a roundabout. 
    Experiments show that LVM has a more stable training process and better control performance compared with Dreamer and SLAC. 
    We conclude that LVM is a promising method to develop an image-input autonomous driving system.
    
    In the future, the potential of LVM in different driving tasks and more complex scenarios will be investigated. Besides, a better latent environment model structure and learning method will be explored.


\bibliographystyle{IEEEtran}
\bibliography{ref}

\end{document}